\documentclass{article} %
\usepackage{arxiv/iclr_sty_for_arxiv,times}
\usepackage{graphicx}

\usepackage{amsmath,amsfonts,bm}

\def\eqref#1{equation~\ref{#1}}

\def\1{\bm{1}}

\DeclareMathAlphabet{\mathsfit}{\encodingdefault}{\sfdefault}{m}{sl}
\SetMathAlphabet{\mathsfit}{bold}{\encodingdefault}{\sfdefault}{bx}{n}

\newcommand{\E}{\mathbb{E}}

\newcommand{\KL}{D_{\mathrm{KL}}}

\DeclareMathOperator*{\argmax}{arg\,max}

\iclrfinalcopy
\usepackage{hyperref}
\usepackage{url}
\usepackage{caption}
\usepackage{subcaption}
\usepackage{xcolor}
\usepackage{multirow}
\usepackage{tikz}
\usepackage{xspace}
\usepackage{bbm}

\newif\ifshowcomments
\showcommentsfalse

\newcommand{\rebuttal}[1]{{#1}}

\newcommand{\method}{Replay across Experiments\xspace}
\newcommand{\met}{RaE\xspace}

\title{%
Replay across Experiments: \\ A Natural Extension of Off-Policy RL}

\author{
Dhruva Tirumala$^{1,2}$ \\
\texttt{dhruvat@google.com} \\
\And
Thomas Lampe$^{1}$
\And 
Jose Enrique Chen$^{1}$
\And 
Tuomas Haarnoja$^{1}$
\And 
Sandy Huang$^{1}$
\And 
Guy Lever$^{1}$
\And 
Ben Moran$^{1}$
\And 
Tim Hertweck$^{1}$
\And 
Leonard Hasenclever$^{1}$
\And 
Martin Riedmiller$^{1}$
\\ $^{1}$Google DeepMind 
\\ $^{2}$University College London (UCL) \\
\And 
Nicolas Heess$^{1}$
\And 
Markus Wulfmeier$^{1}$
}

\begin{document}

\maketitle

\begin{abstract}

Replaying data is a principal mechanism underlying the stability and data efficiency of off-policy reinforcement learning (RL).
We present an effective yet simple framework to extend the use of replays across multiple experiments, minimally adapting the RL workflow for sizeable improvements in controller performance and research iteration times.
At its core, \method~(\met) involves reusing experience from previous experiments to improve exploration and bootstrap learning while reducing required changes to a minimum in comparison to prior work. %
We empirically show benefits across a number of RL algorithms and challenging control domains spanning both locomotion and manipulation, including hard exploration tasks from egocentric vision. 
Through comprehensive ablations, we demonstrate  robustness to the quality and amount of data available and various hyperparameter choices. Finally, we discuss how our approach can be applied more broadly across research life cycles and can increase resilience by reloading data across random seeds or hyperparameter variations.

\end{abstract}
\section{Introduction}

In the last few years, reinforcement learning (RL) has transitioned from a topic of academic study to a practical tool for the generation of controllers across various real-world applications \citep{degrave2022magnetic,Bellemare2020AutonomousNO,47493,9811912}.
These advancements have been enabled, in part, by recent advancements in algorithms improving data efficiency, robustness, and controller performance \citep{haarnoja2019soft, lillicrap2016continuous, schwarzer23abbf}. Nevertheless, many problems remain hard to solve with RL. For instance, high-dimensional and partial observations (e.g. egocentric camera views), high-dimensional action spaces, or reward functions that provide inadequate learning signal can all lead to poor asymptotic performance, high variance, low data efficiency, and long training times. 

A major challenge in RL remains the interaction between data collection and learning. Unlike in the case of supervised learning where all data is available and static, in online RL both the quality and quantity of data available for policy or value function learning changes over the course of an experiment. The difficulty of collecting suitable data, and the complex interactions between data collection and function optimization leads to various problems including failure to learn, instabilities, and the premature convergence of function approximators early in learning  \citep{Nikishin2022ThePB,ash2020warm}.

Experience replay \citep{Lin92} has become a popular component of most, modern off-policy RL algorithms. Storing data collected over the course of an experiment and continuously training policy and value functions with this growing dataset can greatly increase data efficiency and stability of RL algorithms. It does not, however, address issues around premature convergence of the function approximators \citep{ash2020warm}, nor does it directly provide a solution to reusing data from prior experiments.
In this paper, we investigate the possibility of extending the use of replay and reusing data in an iterative setting. Our main insight is that a minimal change to the RL workflow can greatly improve the asymptotic performance of off-policy reinforcement learning algorithms. By reusing interaction data from prior training runs it can further reduce the overall experiment time and thus speed up research iterations. 

A number of studies have previously investigated how prior data (including expert data) can be used to kick-start RL training \citep{vecerik2017leveraging, singh2020cog, nair2020accelerating, Walke2022DontSF, ball2023efficient}. We find that the simplest approach, mixing prior and online data with a fixed ratio, is particularly effective across a wide range of application scenarios and algorithms. Our method dubbed \method (\met) performs as well as or better than alternative approaches with fewer hyperparameter choices to be made. Furthermore, for domains that are normally hard to solve even with state-of-the-art algorithms, we find that it can be advantageous to perform multiple training iterations giving each iteration access to all data from prior training runs, effectively realising a minimalist perspective to lifelong learning. 
We hope that \met's simplicity will enable straightforward integration into existing infrastructure and help improve the efficiency of RL workflows.

The main contributions of our work are as follows:
\begin{itemize}
    \item Introduce and empirically validate our simple strategy to solve challenging tasks. Demonstrate state-of-the art performance on a number of domains including when reusing data from the publicly available offline RL Unplugged benchmark \citep{Gulcehre_2020}.
    \item Demonstrate that our approach works across multiple algorithms including DMPO, D4PG, CRR and SAC-X.
    \item Compare \met to common baselines and recent state-of-the-art methods. Highlight the combination of factors that leads to the effectiveness of our approach.
    \item Provide additional ablations to highlight the robustness of \met to quantity and quality of data collected and hyper-parameters.
\end{itemize}

\section{Method}
\subsection{Background}

We consider the reinforcement learning (RL) problem in which an agent observes the environment, takes an action, the environment changes its state and the agent receives a reward in response. The environment is modeled as a Markov Decision Process (MDP) consisting of the state space $S$, the action space $A$, the transition probability $p(s_{t+1}|s_t, a_t)$ and reward $r(s_t, a_t)$ when taking action $a_t$ in state $s_t$. The agent's behavior is specified using a deep neural network policy $\pi(a_t|s_t; \theta)$ with parameters $\theta$; which we will denote as $\pi(a_t|s_t)$ for brevity.

We optimize the agent to maximize the sum of discounted future rewards, as denoted by:
\begin{equation}\label{eq:rl}
    J(\pi) = \mathbb{E}_{\rho_0\left(s_0\right),p(s_{t+1}|s_t,a_t),\pi(a_t|s_t)}\Big[\sum_{t=0}^{\infty} \gamma^{t} r_t \Big],
\end{equation}
where $\gamma \in [0, 1]$ is the discount factor, $r_t=r\left(s_{t}, a_{t}\right)$ is the reward and $\rho_0\left(s_0\right)$ is the initial state distribution. 

Given a policy $\pi$ , the state-action value function, or critic, $Q(s_t, a_t)$ is defined as the expected discounted return when taking an action $a_t$ in state $s_t$ and then following the policy. %
\begin{align} \label{eq:q_function}
    Q(s_t, a_t) = r(s_t, a_t) + \gamma \mathbb{E}_{p(s_{t+1}|s_t, a_t), \pi(a|s_t)} [ Q(s_{t+1}, a) ].
\end{align}

Modern off-policy algorithms \citep{haarnoja2019soft, barth-maron2018distributional, abdolmaleki2018maximum} usually operate through simultaneous or alternating optimization of policy and state-action value function. In this context, \emph{experience replay} \citep{Lin92} is a fundamental mechanism to decouple data collection from policy and value-function optimization. Data is collected by one or multiple policies (which are commonly obtained over the course of a single experiment) and is stored in a replay buffer from which it can be retrieved to compute updates to the policy or value function.
This approach has multiple benefits including data efficiency, reduced variance of the updates and smoothing the learning process \citep{mnih2015human}.

The success of off-policy algorithms and the need for data-efficient and safe learning schemes has led to growing interest in approaches that allow to reuse data across experiments. One end of a spectrum aims to learn policies entirely offline from a pre-generated, fixed corpus of training data \citep{siegel2020keep, wang2020critic, kumar2020conservative, singh2020parrot, ajay2021opal}. This usually requires specialist algorithmic modifications to avoid instabilities when no online data is available. %
It is often desirable to combine prior data with new data gathered during a learning experiment. 
In the simplest case a policy (and value function) pretrained offline can be then finetuned online, on the same or a related task. While various offline-RL algorithms have been optimised for this setting, simpler off-policy approaches can work well with sufficiently diverse data \citep{yarats2022don}.

The second perspective to re-using data is the straightforward reloading into the replay buffer during online learning. This approach has been successful in bootstrapping learning on tasks with expert demonstrations \citep{vecerik2017leveraging, nair2018overcoming, Walke2022DontSF} or to transfer to new tasks by incorporating data from scripted controllers \citep{singh2020cog, Smith2023LearningAA,jeong2020learning}. 
Finetuning offline learned agents and mixing data into a replay is further combined in the AWAC algorithm \citep{nair2020accelerating}. 
It operates in two phases: in the first phase learning is performed entirely offline by pre-loading the replay buffer with offline data. Then, after a specified number of training steps, online data is mixed into the replay and learning proceeds off-policy. 
AWAC proposes a specific algorithm for both phases that is similar to the formulation of the offline-RL algorithm CRR \citep{wang2020critic}.

While these methods have shown promising gains in data efficiency and performance, they each come with specific assumptions that increase complexity, implementation effort, and restrict their application. These include the requirement for multiple training stages \citep{singh2020cog, Smith2023LearningAA}, the introduction of additional training losses \citep{vecerik2017leveraging, nair2018overcoming,song2022hybrid} or validation with specific architectures and algorithms \citep{nair2020accelerating, Walke2022DontSF,ball2023efficient} which typically require domain-specific hyperparameter tuning. 
The core idea that we are exploring in this paper is that the reuse of prior data can be implemented in the context of most contemporary off-policy RL workflows in a very simple way that is robust and leads to excellent results, without many of the algorithmic complexities or hyper-parameter choices of prior work.

\subsection{\method} 

\begin{figure}[t]
\centering
\includegraphics[width=.8\textwidth]{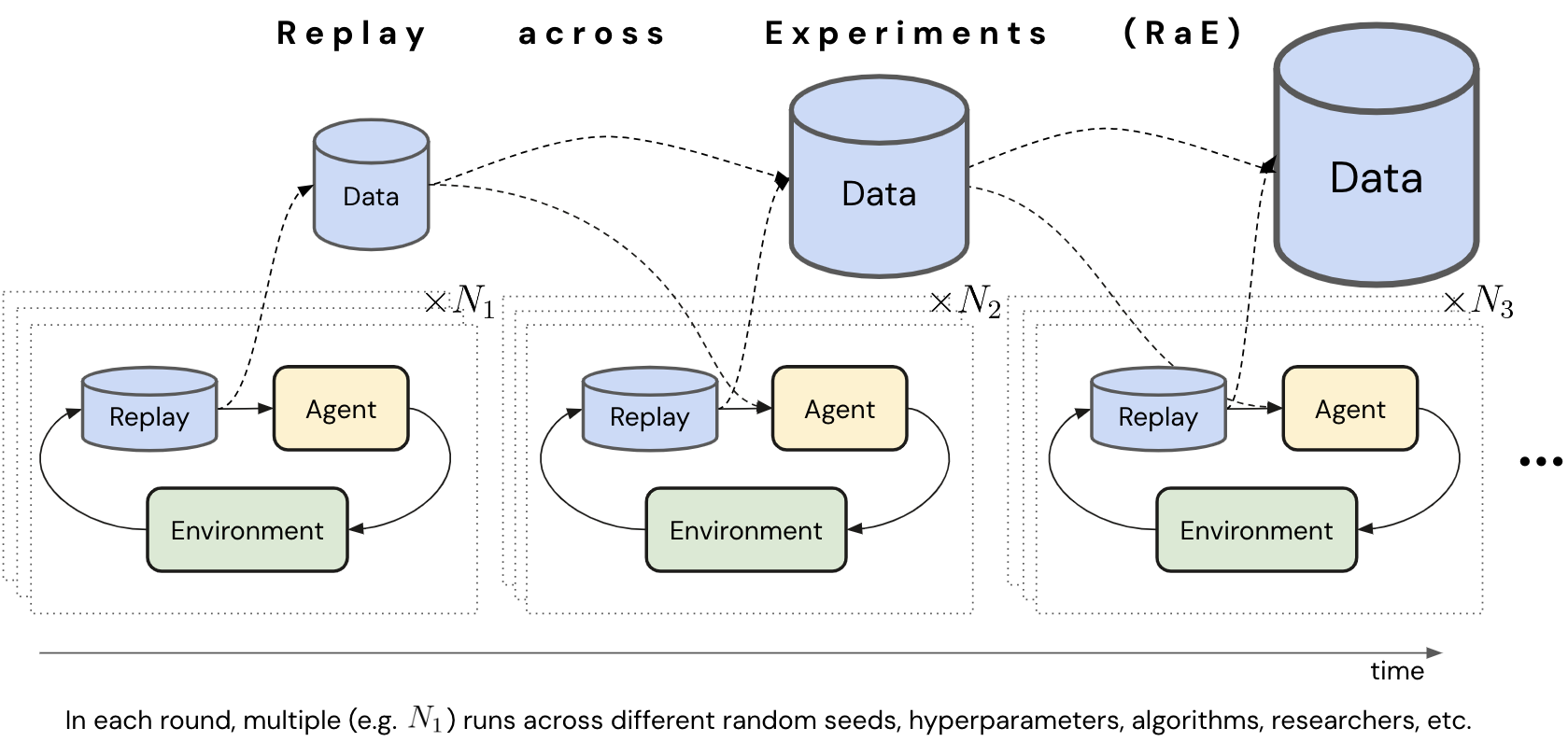
}
\caption{Intuitive schema for the use of regular replay in off-policy RL and the extended application in \met. In every experiment, the agent can learn from a mixture of data from all previous experiments to accelerate learning and increase asymptotic performance.}
\label{fig:rae}
\end{figure}

The basic insight of this work is that reusing data across multiple experiments in off-policy learning is a very simple but effective way to accelerate training and improve final performance. Our key argument is the extension of this insight across all experimentation during a project. Prior data can be used to bootstrap new training runs, and it can further be beneficial to train difficult tasks in multiple iterations, bootstrapping later iterations with the data collected during earlier ones.

This idea is illustrated in Figure \ref{fig:rae}. Whereas off-policy RL is generally focused on reusing data during a single experiment, we consider sequences of experiments where data from earlier training runs is made available during later ones\footnote{We store and reuse all training data across experiments, not just trajectories left in the final replay.}. At the beginning of each training run, policy and value-function are re-initialized in line with stand-alone experiments. The only required algorithmic change is the availability of a second replay mechanism that allows replaying prior and online data with a particular fixed ratio \rebuttal{throughout the course of training} (we use a naive 50/50 mix of offline and online data for our main results, without optimizing this ratio).

This simple approach performs very well without introducing additional algorithmic complexities or the need for hyper-parameter tuning. By learning in the mixed online setting, the agent controls the data distribution and in line with recent results in offline RL \citep{yarats2022don,Lambert2022TheCO}, we show that given the right data distribution, existing off-policy algorithms are sufficient for effectively using this offline data.
This workflow does not require any changes with respect to the RL agent itself and is generally agnostic to agent and architecture changes across experiments. In order to emphasize this aspect, for the main results presented in Section \ref{sec:main_results}, we integrate \met into existing algorithmic workflows across all of the domains considered. Consequently, we evaluate \met across a set of state-of-the-art agents including DMPO \citep{abdolmaleki2018maximum}, SAC-Q \citep{riedmiller2018learning}, CRR \citep{wang2020critic} and D4PG \citep{barth-maron2018distributional} across a number of complex domains.

\section{Experiments}\label{sec:exps}
In this section, we evaluate the performance of \met across feature-based and vision-based, simulated robot locomotion and manipulation settings and standardised RL control benchmarks.

\subsection{Domains}
We consider the following set of challenging domains (visualised in Figure \ref{fig:domains}):

\begin{figure*}[t!]
  \centering
    \includegraphics[width=\textwidth]{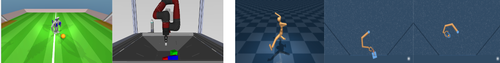}
\caption{\textbf{Domains:} (from left to right) Locomotion, Manipulation stacking and the RL Unplugged domains - Humanoid run, Manipulator insert peg, Manipulator insert ball. \label{fig:domains}}

\end{figure*}
\paragraph{Locomotion Soccer} 
For locomotion, we use a simulated robot soccer task \citep{Haarnoja2023LearningAS} where a simulated OP3 humanoid robot ~\citep{robotisOP3} is rewarded for scoring goals (along with additional shaping rewards) against an opponent. We consider two versions of this task: using proprioception and task-specific features (Locomotion Soccer State) and using proprioception with egocentric visual features (Locomotion Soccer Vision). This second variant is particularly challenging since the agent must learn to score from a sparse reward signal and visual features that make the environment partially observable. Further information on this domain can be found in Appendix \ref{app:soccer}. We use MPO \citep{abdolmaleki2018maximum} with a distributional critic \citep{bellemare2017distributional, barth-maron2018distributional} as underlying off-policy RL algorithm. For our experiments we gather training data of 4e5 and 2e5 episodes \rebuttal{\footnote{Throughout the text we use the scientific notation shorthand where 4e5 refers to $4*10^5$.}} for state and vision respectively. 

\paragraph{Manipulation RGB Stacking} For manipulation, we consider the task of stacking parameterized color-coded objects from \cite{lee2021beyond}. This task builds on visual inputs and only sparse rewards. We use the multi-task SAC-Q off-policy RL algorithm \citep{riedmiller2018learning} in this domain. Further information can be found in Appendix \ref{app:manipulation}.
The offline dataset here consists of 15e4 episodes of training data from a prior experiment with the same algorithm. 

\paragraph{RL Unplugged}
Finally, we evaluate with the offline RL benchmark and dataset RL Unplugged \citep{Gulcehre_2020}, which includes offline data on various simulated control domains. Within these, we focus on the three most challenging Control Suite domains \citep{tassa2020dmcontrol}: humanoid run, manipulator insert peg and manipulator insert ball. The benchmark dataset contains 3000 episodes for humanoid run and 1500 episodes each for the manipulator domains. We use \met with the offline RL algorithm CRR \citep{wang2020critic}, which to our knowledge has state-of-the-art performance for pure offline learning on this benchmark.

The offline data in this setting is notably different from the other domains. In order to collect the data, the authors follow a sub-sampling criterion (see \cite{Gulcehre_2020} for details) to select a subset of episodes which are then randomly stored as single-step trajectories. This is in contrast to our simple protocol of collecting and reusing all data across training with a single algorithm. We include this domain to illustrate the robustness and flexibility of \met as well as to simplify future extensions and comparisons. More details on all domains are presented in Appendix \ref{app:environment}.

\subsection{Baselines}
We compare \met against a number of strong baselines to illustrate its effectiveness.

\paragraph{Fine-tuning}
We first train a policy and state-action value function offline using Critic Regularized Regression (CRR) \citep{wang2020critic} to retain the flexibility of architecture and related changes across experiments. We evaluate a pure behavior cloning agent (BC) and two variants of CRR, CRR binary and CRR exp (see Appendix \ref{app:algorithms}). Subsequently, we choose the best performing seed from across these runs and fine-tune the agent (both policy and value function) online.

\paragraph{AWAC}
AWAC \citep{nair2020accelerating} starts by learning entirely offline for a fixed number of pre-training steps before switching to online iteration. The main difference to finetuning applies to the online phase, where it uses a offline/online shared replay buffer with a specific algorithmic formulation. The authors of this work prescribe 25e3 pretraining steps offline before incorporating online data. For our sweep we also considered 5e4 steps and 1e5 pre-training steps. We also sweep over different values for the Lagrange multiplier $\lambda$ (0.3 and 1) as proposed by the original work.

\paragraph{Random Weight Resetting}
Reloading data for an experiment restart implicitly involves resetting network weights. \cite{Nikishin2022ThePB} observed that network resets on their own can be beneficial for learning. \footnote{\rebuttal{When using RaE, the data distribution throughout learning uses a fixed ratio between offline and online data, in addition to implicitly resetting network weights.}} To tease apart the effect of weight resets from the benefit of data reloading, we consider a baseline where all weights (policy, critic and optimizer) are reset every $K$ policy updates; where $K$ is set to the number of updates used to train the policy that generated the data. We sweep over reset frequencies of $K$, $K/10$ and $K/100$\footnote{Since the data collection methodology for the RL Unplugged domains is different and the number of updates unknown, we instead chose to sweep over values of 1e4, 1e5 and 1e6 updates for these domains}. All experiments are run till convergence or at least $2*K$ updates.

For each method, we report results of the best performing hyperparameter variant averaged across 5 seeds. Importantly, note that \met did not require any tuning and we report results using the same hyperparameter set across all of the main results (50\% offline data). While minor performance improvements with different parameters for \met can be observed in extreme settings (see Table \ref{table:ablation_data_mix}), we find \met has very low sensitivity to hyperparameter choices across a range of diverse domains.

\rebuttal{
\begin{figure*}[t]
\centering
\includegraphics[height=0.25cm,trim={10mm 10mm 10mm 10mm}, clip=true]{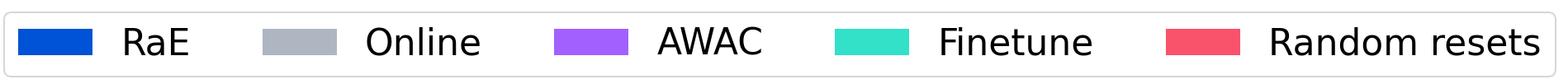}\\
\centering
  \begin{subfigure}[b]{0.2\textwidth}
    \centering
    \includegraphics[width=\textwidth,]{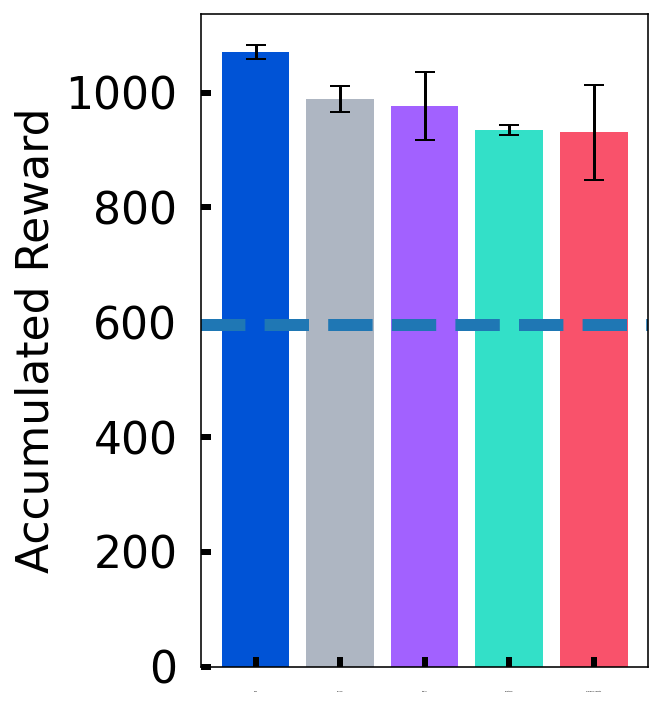}
    \caption{L: Soccer State}
  \end{subfigure}%
  \hfill
  \begin{subfigure}[b]{0.2\textwidth}
    \centering
    \includegraphics[width=\linewidth,]{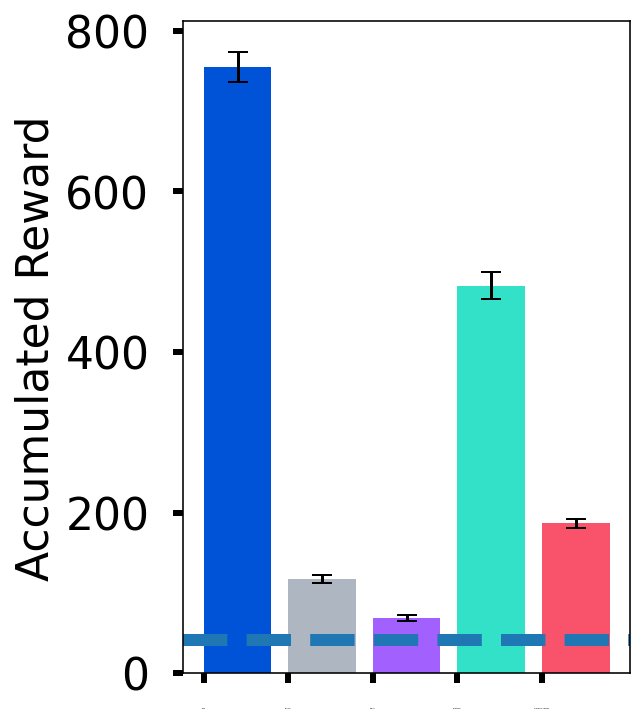}
    \caption{L: Soccer Vision}
  \end{subfigure}%
  \hfill
  \begin{subfigure}[b]{0.2\textwidth}
    \centering
    \includegraphics[width=\linewidth,]{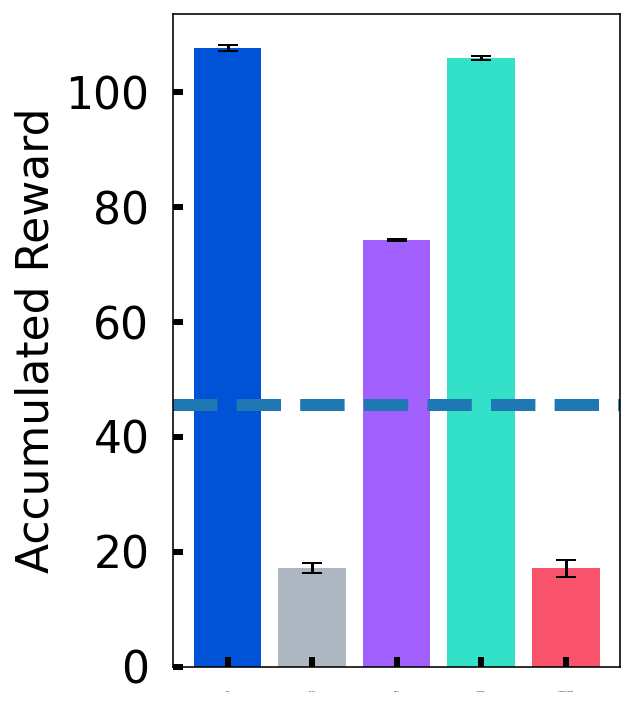}
    \caption{M: Place}
  \end{subfigure}%
  \hfill
  \begin{subfigure}[b]{0.19\textwidth}
    \centering
    \includegraphics[width=\linewidth,]{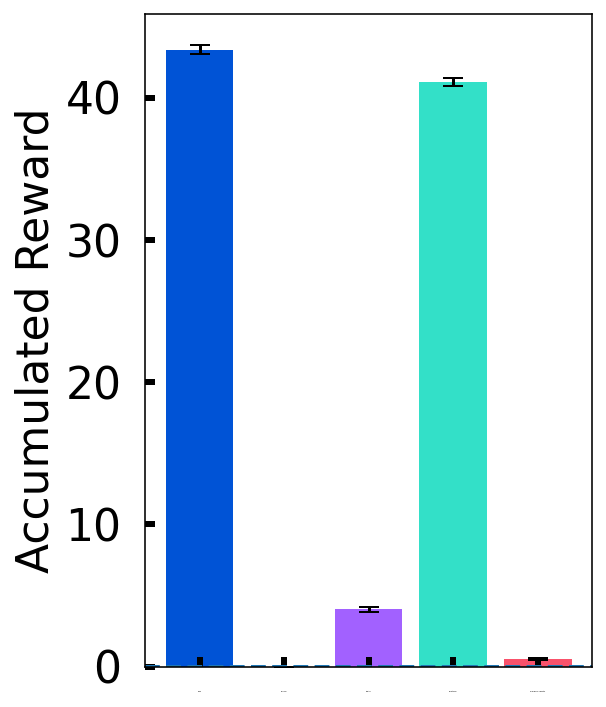}
    \caption{M: Stack Leave}
  \end{subfigure}%
  \hfill
 \vspace{-0.5em}
\caption{Comparison of \met against baselines on the Locomotion Soccer and Manipulation domains (denoted by L: and M: respectively). Y-axis shows accumulated reward \rebuttal{(undiscounted episode return)} for each method; dotted blue line indicates offline learning performance with CRR. \met consistently performs at-par or better than the baselines with a notable improvement on tasks involving vision.}
\label{fig:loco_man}
\end{figure*}
}

\subsection{Main Results}
\label{sec:main_results}

Figure \ref{fig:loco_man} visualises our results on the Locomotion Soccer and Manipulation domains. It compares asymptotic performance achieved by \met and other baselines after convergence as a bar plot with the dark solid line showing \rebuttal{a 95\% confidence interval around the mean averaged across 5 seeds after smoothing over 1000 episodes.}
As the figure shows, \met consistently achieves the highest asymptotic performance across these tasks with a particularly notable improvement on the challenging vision-based `Locomotion Soccer Vision' domain. In the Manipulation settings, both finetuning and \met perform similarly, although finetuning requires choosing the right algorithm and hyperparameter for offline learning.

\begin{figure*}[t]
\centering
\includegraphics[height=0.7cm, clip=true]{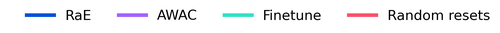}
  \centering
  \begin{subfigure}[b]{0.3\textwidth}
    \centering
    \includegraphics[width=\textwidth,]{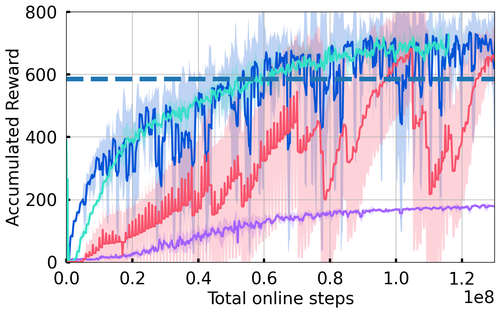}
    \vspace{-1.5em}
    \caption{Humanoid Run}
  \end{subfigure}%
  \hfill
  \begin{subfigure}[b]{0.3\textwidth}
    \centering
    \includegraphics[width=\linewidth,]{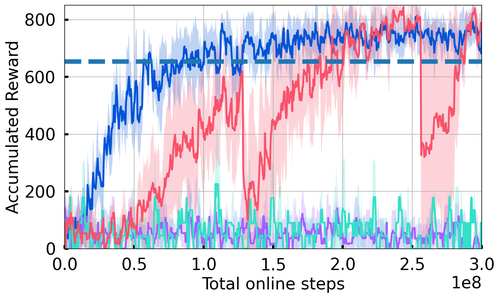}
    \vspace{-1.5em}
    \caption{Manipulator: Insert peg}
  \end{subfigure}%
  \hfill
  \begin{subfigure}[b]{0.3\textwidth}
    \centering
    \includegraphics[width=\linewidth,]{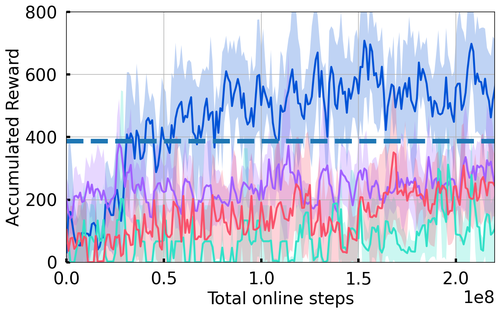}
    \vspace{-1.5em}
    \caption{Manipulator: Insert ball}
  \end{subfigure}%
 \vspace{-0.5em}
\caption{Performance of various mechanisms to reuse offline data using the RL Unplugged dataset. Accumulated rewards \rebuttal{(undiscounted episode return)} on the Y-axis are plotted against the total steps. Each curve plots the mean performance across 5 seeds with standard deviation shown as shaded regions. \met consistently performs the best across all tasks.}
\label{fig:rl_unplugged}
\end{figure*}

Figure \ref{fig:rl_unplugged} compares \met with all baselines that can use the RL Unplugged offline dataset. The figure shows the accumulated reward over the total number of additional \textit{online} data consumed for each method. The performance of the best pure offline variant (CRR) is shown as a dotted blue line. 
We observe that \met performs at-par or better than other comparable methods with the biggest difference being in the manipulator tasks. \rebuttal{Methods that use offline learning (CRR and finetuning) work well on the densely rewarded Humanoid run domain but peform poorly in the sparsely rewarded manipulator domains whereas \met works well across all settings.}

\subsection{Ablations}
\label{sec:ablations}

In this section, we present a set of experiments designed to answer the following questions:%
\begin{itemize}
    \item How much data is required for performance improvements?
    \item What kind of data is best for mixing: expert data, early training data or a mix?
    \item How sensitive is \met to the ratio of online to offline data in different data regimes?
    \item Is there an advantage of applying \met repeatedly across experiment iterations?
    \item Is \met agnostic to the underlying choice of algorithm?
\end{itemize}

Unless otherwise specified, we present results on the `Locomotion Soccer State` task with the DMPO algorithm.

\paragraph{Analysis with varying data}
We answer the first three of these questions together in this section. For this analysis we use multiple subsets of the data collected for the `Locomotion Soccer State' task, divided into three regimes:
\begin{itemize}
    \item \textbf{High Return} We consider data generated only from the end of training \rebuttal{from a complete training run}. This corresponds to highly rewarding trajectories or `expert' data.
    \item \textbf{Mixed Return} We consider data sampled uniformly at random throughout learning. This corresponds to a mixed regime with of high and low return trajectories.
    \item \textbf{Low Return} Finally, we consider data generated only from the start of training. This corresponds to early low return data.
\end{itemize}

 \rebuttal{High and Low Return data are sampled according to recency while Mixed return samples data uniformly.}
For each regime, we consider 2 datasets: one with 1e5 episodes\footnote{The original dataset consists of 4e5 episodes.} and another with just 1e4 episodes. \rebuttal{The purpose of this analysis is to understand how best to adapt \met when in a regime with limited data.} For each data subset we then consider data mixing at different ratios of online to offline data: in addition to the 50\% mix that was used in the main results we consider mixes of 70, 80 and 90\% online to offline data. For each setting, we consider the asymptotic performance reached when using \met on the `Locomotion Soccer State` task.

Table \ref{table:ablation_data_mix} shows the final asymptotic reward as a percentage of the \textit{final} reward achieved when learning online from scratch, which corresponds to 4e5 episodes. To improve the ease of interpretation the cells are colored based on the reward achieved: yellow for below 90\% performance, orange for between 90-100\% and blue for greater than 100\%. 

A few interesting patterns emerge which we summarize below:
\begin{itemize}
    \item In the lower data regime (1e4 episodes) a mixture with more online data is beneficial. However, as more data becomes available, a lower ratio works better (of 70-80\% at 1e5 episodes). \rebuttal{We hypothesize that using more online data for learning prevents over-fitting to a small set of offline trajectories.}
    \item In the lower data regime, low return data tends to be the most beneficial. As the dataset size increases though, a mix of high and low return trajectories provide a greater benefit. Surprisingly, expert data is the least beneficial in both regimes. This indicates that the advantage of mixing data may stem from the benefit of having a larger state distribution with mixed rewards.
    \item Gains in performance can be achieved with as little as 1e4 episodes. This is particularly promising since even a small amount of prior data can considerably improve results.
\end{itemize}

\begin{table}
\centering
\begin{tabular}{||p{2.5cm} | p{1.2cm} | p{1.2cm} || p{1.2cm} | p{1.2cm} || p{1.2cm} | p{1.2cm} ||}
\hline
\hline
\multicolumn{1}{|c|}{} &
\multicolumn{2}{|c|}{\textbf{High return}} & \multicolumn{2}{|c|}{\textbf{Mixed return}} & \multicolumn{2}{|c|}{\textbf{Low return}} \\
\hline
\hline
\tikz{\node[below left, inner sep=1pt] (def) {Data mix};%
      \node[above right,inner sep=1pt] (abc) {Episodes};%
      \draw (def.north west|-abc.north west) -- (def.south east-|abc.south east);} &
\textbf{10,000} & \textbf{100,000}  &
\textbf{10,000} & \textbf{100,000}  &
\textbf{10,000} & \textbf{100,000}  \\
\hline
50\% Online & \textcolor{red}{51\%} & \textcolor{orange}{90\%} & \textcolor{red}{80\%} & \textcolor{blue}{119\%} &  \textcolor{orange}{97\%} & \textcolor{blue}{112\%} \\
\hline
70\% Online & \textcolor{red}{80\%} & \textcolor{blue}{110\%} & \textcolor{blue}{101\%} & \textcolor{blue}{124\%} &  \textcolor{blue}{108\%} & \textcolor{blue}{121\%} \\
\hline
80\% Online & \textcolor{orange}{98\%} & \textcolor{blue}{118\%} & \textcolor{blue}{106\%} & \textcolor{blue}{126\%} &  \textcolor{blue}{110\%} & \textcolor{blue}{123\%} \\
\hline
90\% Online & \textcolor{blue}{108\%} & \textcolor{blue}{120\%} & \textcolor{blue}{113\%} & \textcolor{blue}{120\%} &  \textcolor{blue}{108\%} & \textcolor{blue}{121\%} \\
\hline
\hline
\end{tabular}
\caption{Performance comparison when using \met in different data regimes with various mixes of online and offline data. Each cell represents asymptotic performance improvement as a percentage of the performance reached when learning from scratch \textit{to convergence}\rebuttal{(100\% online)}. Colors are used to indicate the trend: red for below 90\% performance, orange between 90 to 100\% and blue when greater than 100\%}
\label{table:ablation_data_mix}
\end{table}

\begin{figure*}[t]
\includegraphics[height=0.7cm, clip=true]{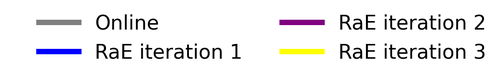}
\hfill
\includegraphics[height=0.65cm, clip=true]{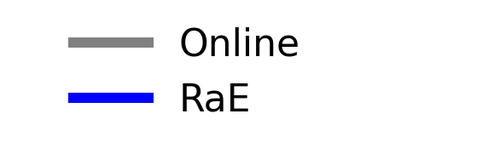}
\hfill
\includegraphics[height=0.65cm, clip=true]{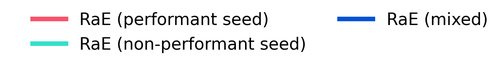}
  \centering
  \begin{subfigure}[b]{0.33\textwidth}
    \centering
    \includegraphics[width=\textwidth,]{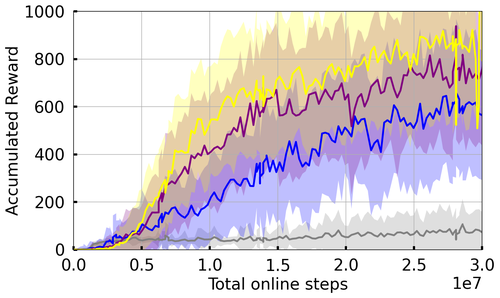}
    \vspace{-1.5em}
    \caption{Multiple iterations of RaE}
    \label{fig:ablation_reuse_rerun}
  \end{subfigure}%
  \hfill
\begin{subfigure}[b]{0.33\textwidth}
    \centering
    \includegraphics[width=\textwidth,]{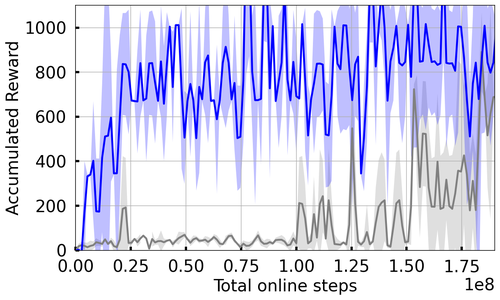}
    \vspace{-1.5em}
    \caption{RaE with D4PG}
    \label{fig:ablation_d4pg}
  \end{subfigure}%
  \hfill
 \begin{subfigure}[b]{0.33\textwidth}
    \centering
    \includegraphics[width=\textwidth,]{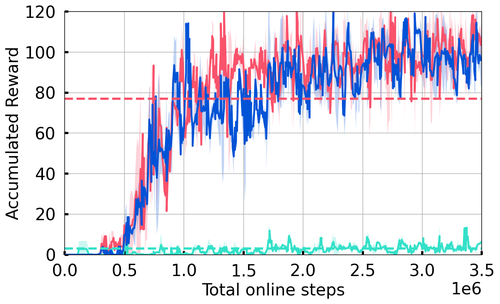}
    \vspace{-1.5em}
    \caption{Reloading across seeds}
    \label{fig:ablation_seeds}
  \end{subfigure}%
\caption{a) \met can be applied iteratively on the `Locomotion Soccer State' domain. In some cases it may be preferable to have multiple smaller iterations although a performance plateau may be reached eventually. b) \met on the `Locomotion Soccer State' domain continues to improve performance with the D4PG algorithm. c) Combining data from seeds of a high variance experiment. \met performs well even if just a single original seed achieves high reward. Performance of the original data generating experiments plotted as dotted lines. All figures plot mean accumulated reward against total online steps with standard deviation shown as a shaded region.}
\end{figure*}

\paragraph{Iterative improvement}
Figure \ref{fig:ablation_reuse_rerun} shows the improvement in performance when iteratively applying \met on the `Locomotion Soccer State' domain. For this experiment, we begin by generating a smaller dataset of 1e4 episodes and then apply \met to that. We collect the data from this already improved run (\met iteration 1) and apply \met again. We observe that there is a small gain in asymptotic performance and speed of learning even on the second iteration although a performance plateau is reached on a third iteration. This result suggests that in some settings it may be preferable to break a single training run into smaller runs for iterative performance improvements.

\paragraph{Indifference to algorithm}
For the main analysis of Section \ref{sec:main_results}, we show that \met can improve performance across a range of domains with different underlying algorithms. Figure \ref{fig:ablation_d4pg} reinforces this point on the `Locomotion Soccer State' domain where we show a considerable gain in performance when applying \met using the D4PG \citep{barth-maron2018distributional} algorithm for both data collection and mixing. 

\paragraph{Increased robustness}
By applying \met to combine data across seeds and hyperparameters from previous experiments we can add crucial robustness to the underlying agent's variance in performance. Figure \ref{fig:ablation_seeds} demonstrates robust performance when reloading data across random seeds in comparison to purely high or low performing data sources.

\section{Related Work}
The most common paradigms to collect and reuse data in RL involve either offline agent datasets (e.g. RL Unplugged \citep{Gulcehre_2020} and D4RL \citep{Fu_2020}) or the use of human or expert demonstration trajectories \citep{vecerik2017leveraging, Bohez2022ImitateAR}. While some work has focused on solving benchmark challenges \citep{wang2020critic, kumar2020conservative} others have focused on improving online performance by integrating offline data. For example, \cite{singh2020parrot} and \cite{ajay2021opal} learn hierarchical architectures offline and reuse them to solve more challenging tasks online. Related to our approach, \cite{Smith2023LearningAA} mix experience using controllers designed for a different setting with online data to improve learning efficiency for quadruped robots. Our work instead highlights the merits of directly transferring data to improve asymptotic performance on a single domain.

\paragraph{Learning from Data Combinations}
The idea of mixing expert data sources to bootstrap learning on the same domain has also been studied in a number of related works \citep{vecerik2017leveraging, nair2018overcoming, singh2020cog,  Walke2022DontSF, Davchev2021WishYW, lee2022offline, ball2023efficient, nair2020accelerating}. While these methods have shown impressive results on a range of domains, the complexity and specificity of various algorithmic assumptions limit their generality. 
While \cite{vecerik2017leveraging} also mix expert demonstration data into the replay, they include a prioritized replay and a mix of 1-step and N-step returns with L2 regularization in their setup. In a similar vein, \cite{nair2018overcoming} introduce a BC-loss, Q-filter and state-reset mechanism as part of their data mixing approach. Other work relies on multi-stage procedures when mixing data: \cite{singh2020cog} distill prior data from skill experts to be later mixed with data from a scripted controller followed by the application of CQL; \cite{Walke2022DontSF} train a multi-task policy with separate forward and backward policies that are optimized separately.
Even work that uses existing offline datasets as opposed to expert demonstrations inevitably include other components that increase their complexity. For instance, \cite{lee2022offline} finetune an ensemble of policies trained offline using a prioritized replay and a density ratio to choose the data mixture to improve performance with the D4RL dataset. On the same domain, \cite{ball2023efficient} begin with a similar formulation to ours but then argue for random ensemble distillation and per-environment design choices and LayerNorm when training the Q-function with which they demonstrate improvements using a single off-policy algorithm. Finally, the AWAC algorithm \citep{nair2020accelerating} that we compare against sits in between finetuning and data mixing but uses a specific algorithmic formulation which introduces a number of domain-dependent hyperparameters. 

In contrast, our focus is on the simplicity of the method: we empirically demonstrate the effectiveness and versatility of mixing previous data with a fixed ratio using many off-policy RL algorithms.

\paragraph{Resetting Network Parameters in Reinforcement Learning}
A consequence of restarting experiments with mixed data is the resetting of neural network parameters. Resetting weights can change the learning dynamics and prevent overfitting which has been shown to be beneficial in both supervised learning \citep{ash2020warm} and RL \citep{Nikishin2022ThePB, schwarzer23abbf}. Under certain settings where parts of a network are reset at specific intervals, these methods have been shown to greatly improve learning efficiency on challenging domains like Atari. In this work, we demonstrate that reusing previous experience can improve asymptotic performance on a range of tasks. Combined with its simplicity, we hope our approach can be included as part of the natural iteration cycle in RL and robotics in particular.

\section{Discussion}

As RL continues to move from the object of study to a practical engineering tool for control \citep{degrave2022magnetic, Bellemare2020AutonomousNO, 47493, 9811912, kaufmann2023champion, osinski2020simulation}, simple and practical methods that can be generally applied will become increasingly important. %

We can apply \met throughout the lifetime of a project, across multiple experiments, algorithms, and hyperparameter settings. In the following paragraphs, we provide some intuition on some practical use-cases for \met. %

\paragraph{Lifelong / Project-long learning}
Research into novel domains nominally involves many iterations of trial-and-error to achieve state-of-the-art performance. While learnings from early iterations of experimentation inform algorithmic choices, the data collected in these trials is commonly discarded and never reused. Our initial analysis in Section \ref{sec:ablations} indicates that even low-return data can be useful to boost performance. With costs of data storage typically being far lower than compute, a different workflow where all experimental data (particularly in domains like robotics) is stored and reused to bootstrap learning could improve efficiency across project lifetimes.  

\paragraph{Multiple Source Experiments}
\met can also be applied in domains with many related source experiments. For example, consider many tasks defined via different reward functions but with the same underlying dynamics (e.g. the family of all manipulation tasks with a specific morphology). High-return data on some tasks may result in lower returns in another. However this data is still informative and may be useful in improving exploration when transferred. 

\paragraph{Multiple Hyperparameters and Seeds} %
Combining data from different variants of the same experiment (such as random seeds or hyperparameters) can enable better use of large, expensive sweeps. When reloading data with high performance variance over these parameters, the new experiment can benefit from the best option as initially discussed in Section \ref{sec:ablations}.

\paragraph{Potential Limitations and Strategies for Mitigation}
Across all experiments, \met demonstrates better or similar performance to state-of-the-art baselines for efficient off-policy RL without requiring per task hyperparameter tuning. However the applicability of the method is limited by the ability to reuse data. For example, changes in dynamics or experimental settings might invalidate previously collected data \footnote{\rebuttal{However in practice, this may not be much of an issue: see Appendix \ref{app:changing_dynamics}}}. In such cases, an intermediate step to collect transitional data between old and new settings might be useful. 
\rebuttal{Another challenge that may arise with widespread use of \met is in the comparison of new algorithms. RL already suffers from issues of reproducibility \citep{henderson2018deep}; if algorithms incorporate \met, subtle changes in the data distribution when learning may create large differences in performance. One solution to this may be to specify deterministic orderings for benchmark datasets using fixed random seeds.}

\section{Conclusions}

With this work, we introduce an effective and, importantly, simple workflow change for off-policy RL experimentation. Reloading data requires minimal infrastructure and can greatly improve performance as shown in Section \ref{sec:exps}. This makes it particularly useful from a project-long (or life-long) learning perspective in RL experimentation. We believe that as our understanding of RL improves and its use as an engineering and control tool becomes more commonplace, simplicity is key to effective integration.

\bibliography{main}
\bibliographystyle{iclr/iclr2024_conference}

\appendix
\section{Environment Details}
\label{app:environment}
\subsection{Locomotion Soccer}\label{app:soccer}
The `Locomotion Soccer' environment introduced by \cite{Haarnoja2023LearningAS} consists of a Robotis OP3~\citep{robotisOP3} robot which is placed in a $4m \times 4m$ walled arena with a soccer ball. The robot must remain in its own half and scores goals when the ball enters a goal area $0.5m times 1m$ positioned across the center of the back wall in the other half. The robot must also contend with a random agent in the other half. The reward for this task is given by:

\begin{equation}
\label{eq:scoring}
    R_{goalscoring} = R_{score} + R_{upright} + R_{maxvelocity}
\end{equation}

where: $R_{score}$ is 1000 on the single timestep where the ball enters the goal region (and then becomes unavailable until the ball has bounced back to the robot's own half); $R_{maxvelocity}$ is the norm of the planar velocity of the robot's feet in the robot's forward direction; $R_{upright}$ is 1.0 when the robot's orientation is close to vertical, decaying to zero outside a margin of $12.5^{\circ}$.  Additionally, episodes are terminated with zero reward if the robot leaves its own half, or body parts other than the feet come within 4cm of the ground. 
The agent chooses a desired joint angle every 25 milliseconds (40Hz) based on input observations that include the joint angles, angular velocity and gravity direction of the torso. 
For `Locomotion Soccer State' the agent also receives the coordinates of the ball, goal and opponent. For the `Locomotion Soccer Vision` domain the robot instead receives a $40 \times 30$ render of the egocentric camera. 

We use the experimental settings and networks described in \cite{Haarnoja2023LearningAS} for this setting. We use a batch size of 128 with 5 seeds for the main results in Section \ref{sec:exps} and a batch size of 256 with 2 seeds for the experiments in Section \ref{sec:ablations}. 
\subsection{Robot Manipulation}\label{app:manipulation}

The RGB Stacking benchmark introduced by \cite{lee2021beyond} consists of parametric extruded shapes, which need to be stacked on top of each other by a fixed robotic manipulator, with the role of objects in the configuration being denoted via color coding.
The benchmark uses a Rethink Sawyer robot arm, controlled in Cartesian velocity space on which a Robotiq 2F-85 parallel gripper is mounted.

We focus on the "skill mastery" challenge of the benchmark as described in \cite{lee2021beyond}, where five fixed triplets of objects are used both for training and testing, with the goal being to place the red object on the blue one, while ignoring the green one. Each object triplet highlights a different manipulation aspect, such as balancing or reorientation.

For the SAC-X setup, we use a curriculum of sub-tasks that can be chosen by the scheduler. These follow the reward terms of the staged reward terms used in \cite{lee2021beyond}, and intuitively contain sub-tasks for reaching, lifting, placing and stacking the red object.
For evaluation, we report the final "stack-leave" reward term, which is a sparse reward given when the red object is placed precisely on the blue one, and the arm has been moved away by a predefined distance.

The observation provided to the agent consists of the images of three static cameras, as well as proprioception data from the robot and gripper, including joint angles, velocities and torques, as well as a simulated force-torque sensor attached to the wrist. It notably does not contain the tracked position of the objects, so the agent must learn to achieve the goal primarily from vision inputs.

In order to solve this task from visual inputs we use a network architecture where both policy and value function use a convolutional neural network with residual blocks (as used in \citep{espeholt2018impala}) that is then passed through an MLP for multi-task output for each of the subtasks. Both networks use (2, 2, 2) residual blocks with convolutional layers with 16, 32 and 32 channels respectively. The output of the policy network is used to condition a multi-headed Gaussian where each Gaussian output represents the policy output for a sub-task. Similarly the value function outputs a multi-headed categorical for each task . Appendix \ref{app:algorithms} describes how this can be used by the SAC-Q algorithm. All experiments are run with 5 seeds for the main results of Section \ref{sec:main_results} and 2 seeds for the ablations in Section \ref{sec:ablations}.
\subsection{RL Unplugged}\label{app:rlunplugged}
We consider 3 datasets from the RL Unplugged benchmark \citep{Gulcehre_2020}: Humanoid run, Manipulator insert peg and Manipulator insert ball; which are categorized as `hard' domains by \cite{wang2020critic}. The RL Unplugged datasets were collected on the DeepMind Control Suite \citep{tassa2018control} implemented in the MuJoCo \citep{todorov2012mujoco} simulation framework. 

Humanoid run consists of 3000 episodes of a 21 dimensional simulated humanoid walker that is rewarded for running forwards while staying upright. The data for this domain is generated using D4PG \citep{barth-maron2018distributional}. The Manipulator insert ball and Manipulator insert peg domains consist of 1500 episodes each where a 5-dimensional manipulator is sparsely rewarded for inserting a ball and peg respectively into a specified hole. Data for this domain was generated using V-MPO \citep{song2020vmpo} since D4PG was unable to solve the tasks. The data generated on these tasks is then reduced in size via sub-sampling and the number of successful episodes in each dataset is reduced by $2/3$ to ensure the data does not contain too many successful trajectories. For all domains each episode consists of 1000 timesteps. 

We use experimental settings as described in \cite{wang2020critic} for this domain and use their networks for the offline and finetuning experiments. For the other baselines and \met~ we use a an MLP architecture with sizes (256, 256, 128) for the policy and (512, 512, 256) for the value function. The policy network output is used to parameterize a Mixture of Gaussian (MoG) with 5 components as in \cite{wang2020critic}. Since the dataset only involves single-step transitions we use 1-step return for all baselines except the Random resets where we found 5-step returns performs significantly better.

\section{Algorithm Details}
\label{app:algo_network_details}
\label{app:algorithms}
In this section we describe the algorithms used in the main text in more detail.

\paragraph{Policy Evaluation}

We use the critic update described in \cite{barth-maron2018distributional} for $N-$step returns:
\begin{equation}
    (\mathcal{T}_{\pi}^{N}Q)(s_0, a_0) = r(s_0, a_0) + \E\left[\sum_{n=1}^{N-1} \gamma^n r(s_n, a_n) + \gamma^{N}Q(s_n, \pi(a_n|s_N))|s_0, a_0\right] \nonumber
\end{equation}
where the expectation is with respect to $N-$step transition dynamics and the distribution $Q$ and $\mathcal{T}$ represents the distributional Bellman operator. For our experiments we parameterize the critic as a categorical distribution with the number of bins set based on the domain. We set $N=5$ for our experiments except in the RL unplugged domains where only single-step trajectories exist in the dataset. However, we use $N=5$ for the random resets baseline in this domain since it can be run online and is not constrained by the limitation in the data.

\paragraph{DMPO}
For our online experiments in `Locomotion Soccer' and `Manipulation Stacking`, we use the Maximum a-posteriori Policy Optimisation algorithm (MPO) \citep{abdolmaleki2018maximum} to adhere to \cite{Haarnoja2023LearningAS}. MPO optimizes the RL objective in an E and M-step. The E-step update optimizes:

\begin{align}
    \max_{q} \int_s \mu(s) \int_a q(a|s)Q(s, a) \,da \,ds  \nonumber \\
    \,s.\,t. \int_s \mu(s) \KL(q(a|s) || \pi(a|s)) \,ds < \epsilon \nonumber 
\end{align}
where $q(a|s)$ an improved non-parametric policy that is optimized for states $\mu(s)$ drawn from the replay buffer. The solution for $q$ is shown to be:
\begin{align}
    q(a|s) &\propto  \pi(a|s, \theta) \exp(\frac{Q_\theta(s, a)}{\eta^*}).\nonumber
\end{align}
where $Q$ is computed as an expectation over the categorical distribution. In the M-step, an improved policy is the obtained via supervised learning:
\begin{align}
    \pi^{n+1} &= \arg\max_{\pi_\theta} \sum_{j}^{M} \sum_{i}^{N} q_{ij} \log \pi_{\theta}(a_i|s_j). \nonumber \\
    &\,s.\,t. \KL(\pi^{n}(a|s_j) || \pi_{\theta}(a|s_j)) \nonumber
\end{align}

\paragraph{SAC-Q}

For the online experiments in 'Manipulation Stacking', we use the Scheduled Auxiliary Control (SAC-X) algorithm for multi-task exploration \citep{riedmiller2018learning}. SAC-X builds on a multi-headed network architecture to represent both critic and policy, with a single "torso" per network shared across multiple tasks, and a separate output "head" for each task. At any given time, the active head to use is decided by a scheduler process. Here, the scheduler itself is also being learned, using the SAC-Q variant of the algorithm. At fixed intervals during each episode, a new sub-task may be chosen, and the scheduler trains a Q-function that optimizes the sequence of tasks executed such that it maximizes the total return of one task designated as the main goal.

To update the policy and Q-function, SAC-Q may use any update rule. Here, we use the MPO algorithm with a distributional critic, as described above.

\paragraph{CRR}
For the offline experiments in Section \ref{sec:main_results} we use the Critic Regularized Regression (CRR) algorithm \citep{wang2020critic} which achieves state-of-the-art performance on the RL Unplugged benchmark. CRR uses the same policy evaluation step described above to learn a distributional critic. However the policy in CRR is trained by filtering data via the Q function by optimizing:
\begin{align}
    \argmax_{\pi} \E_{(s, a)\sim D}\left[f(Q, \pi, s, a)\log\pi(a|s)\right] \nonumber,
\end{align}
where the states and actions are drawn from the data source $D$ and $f$ is a non-negative, scalar function whose value increases monotonically with $Q$. We consider both variants of CRR introduced in the paper defined by choices of $f$:
\begin{align}
    f := \mathbbm{1} \left[ A(s, a) > 0 \right], \label{eq:crr_indicator} \\
    f := \exp \left(A(s, a)/\beta \right) \label{eq:crr_exp},
\end{align}
where $A(s, a)$ is the advantage function which can be computed using:
\begin{align}
    A(s, a) = Q(s, a) - \frac{1}{m} \sum_{j=1}^{m} Q(s, a^j);  a^j \sim \pi(.|s). \nonumber
\end{align}
Following the authors of CRR, we refer to CRR with $f$ from Equation \ref{eq:crr_indicator} as CRR binary and Equation \ref{eq:crr_exp} as CRR exp.

\paragraph{AWAC}
For the AWAC baseline in the main text we follow the description from \cite{nair2020accelerating} where learning proceeds in two stages: entirely offline initially and then with online data added to the same replay buffer. The transition between these two stages is defined by a hyper parameter which is set to 25,000 steps in the original work. For our experiments we sweep over an additional value of 50,000 steps and 100,000 steps for the RL Unplugged domain. 

AWAC also introduces an algorithm which is similar to the CRR exp formulation described above with a slightly different notation that introduces a temperature $\lambda=\frac{1}{\beta}$ above. As per the original work we test the method with $\lambda$ set to 0.3 and 1.

\section{Additional Results}
\label{app:add_results}

\subsection{Manipulation results on remaining tasks}

\begin{figure*}[t]
\centering
\includegraphics[height=0.3cm,trim={10mm 10mm 10mm 10mm}, clip=true]{figures/main/bar_plot_legend.png}
\centering
  \begin{subfigure}[b]{0.2\textwidth}
    \centering
    \includegraphics[width=\textwidth,]{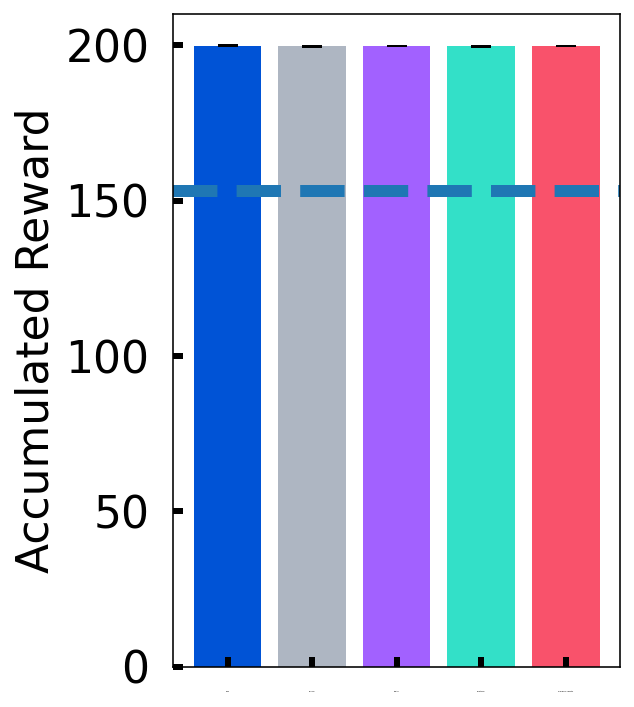}
    \caption{Open}
  \end{subfigure}%
  \hfill
  \begin{subfigure}[b]{0.2\textwidth}
    \centering
    \includegraphics[width=\linewidth,]{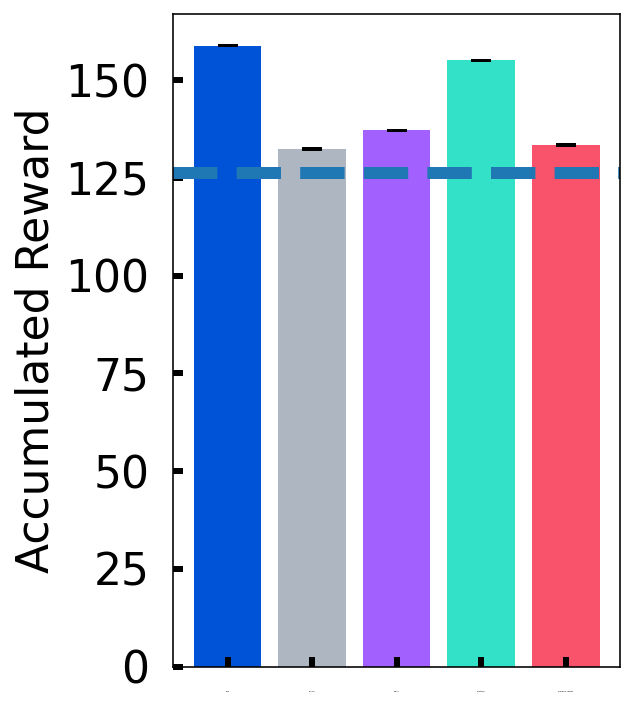}
    \caption{Reach Grasp}
  \end{subfigure}%
  \hfill
  \begin{subfigure}[b]{0.2\textwidth}
    \centering
    \includegraphics[width=\linewidth,]{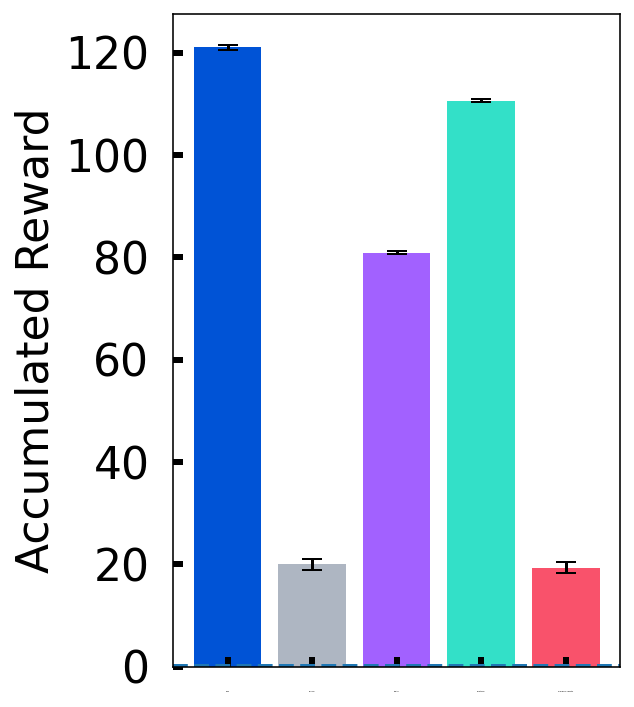}
    \caption{Lift}
  \end{subfigure}%
  \hfill
  \begin{subfigure}[b]{0.2\textwidth}
    \centering
    \includegraphics[width=\linewidth,]{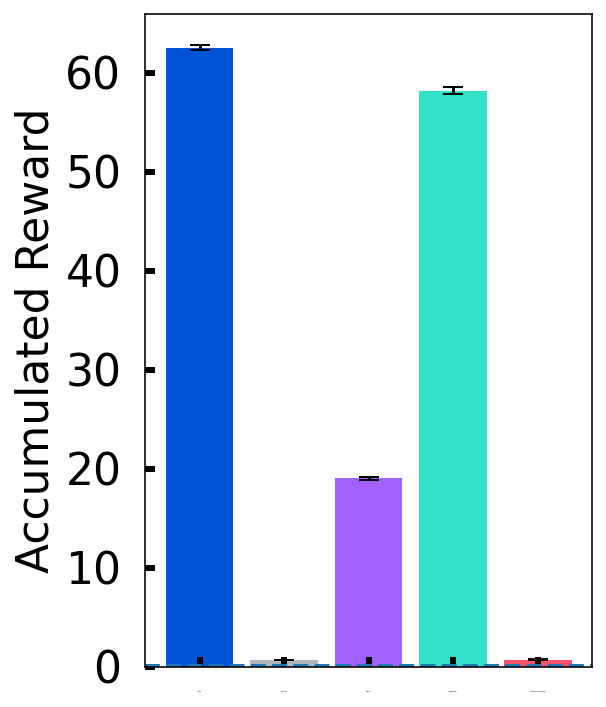}
    \caption{Stack}
  \end{subfigure}%
  \hfill
 \vspace{-0.5em}
\caption{Comparison of \met against baselines on Open, Reach Grasp, Lift and Stack sub tasks in the Manipulation RGB stacking setting. Average performance across 5 seeds is plotted on the Y-axis with standard deviation as as dark lines.}
\label{fig:app_man_remaining}
\end{figure*}
Figure \ref{fig:app_man_remaining} compares \met against the other baselines from Section \ref{sec:main_results} on the remaining sub-tasks in the `Manipulation RGB Stacking' setting. While all methods fare well on the easier tasks like `Open' and `Reach Grasp', \met performs well on the harder `Lift' and `Stack' tasks similar to the trend on `Place' and `Stack Leave' that were presented in the main text.

\subsection{Finetuning with \met}
\begin{figure*}[t]
\centering
\includegraphics[height=.8cm,clip=true]{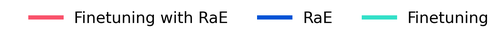}
  \centering
  \begin{subfigure}[b]{0.45\textwidth}
    \centering
    \includegraphics[width=\textwidth,]{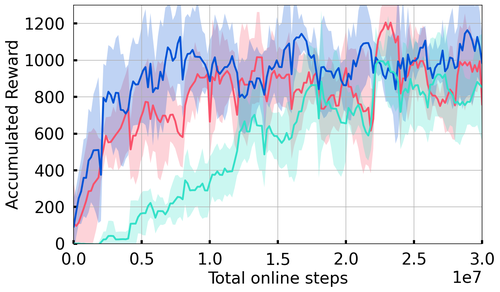}
    \vspace{-1.5em}
    \caption{Locomotion Soccer State}
  \end{subfigure}%
  \hfill
  \begin{subfigure}[b]{0.45\textwidth}
    \centering
    \includegraphics[width=\linewidth,]{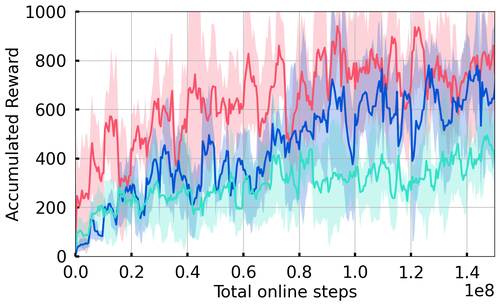}
    \vspace{-1.5em}
    \caption{Locomotion Soccer Vision}
  \end{subfigure}%
 \vspace{-0.5em}
\caption{Comparison of finetuning, \met~ and finetuning with \met~ on the `Locomotion Soccer` state and vision tasks (left and right respectively). Finetuning with \met~ combines the best of both worlds: improved learning speed with high asymptotic performance. Accumulated reward on the Y-axis is plotted against total online steps on the X. Results are averaged across 5 seeds with the shaded region representing the standard deviation.}
\label{fig:ablation_finetune}
\end{figure*}

The `finetuning' baseline considered in the main experiments of Section \ref{sec:main_results} is orthogonal to the core idea of \met~ and as such, we can apply \met~ when finetuning. Figure \ref{fig:ablation_finetune} compares \met~, finetuning and finetuning with \met~ on the `Locomotion Soccer` tasks from state and vision. We observe that finetuning with \met~ performs well across both domains matching the high asymptotic performance of \met~ and learning faster by taking advantage of the network weights pre-trained offline. 

\subsection{Effect of mixing ratio on full dataset}
\rebuttal{
The results presented in Table \ref{table:ablation_data_mix} show the effect of mixing different ratios of data when using \met~ with smaller subsets of data (of 10,000 and 100,000) episodes. Figure \ref{fig:ablation_data_mix_fulldata} instead presents the same analysis when using the full dataset (of 4e5 episodes) on the `Locomotion Soccer State' task. In this setting we see a clearer advantage of using offline data where a mixture of 50 to 70 percent of offline data works better than using more online data (90 \%). In fact, as the figure shows, a mixing ratio of 90 \% may even slightly degrade performance when compared to learning purely online.
}
\subsection{Effect of changing dynamics}
\label{app:changing_dynamics}
\begin{figure*}[t]
\centering
\includegraphics[height=0.7cm,clip=true]{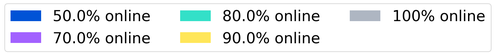}
\includegraphics[height=0.6cm,clip=true]{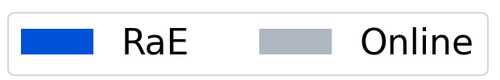}

\begin{subfigure}[b]{0.3\textwidth}
  \centering
  \includegraphics[height=5cm]{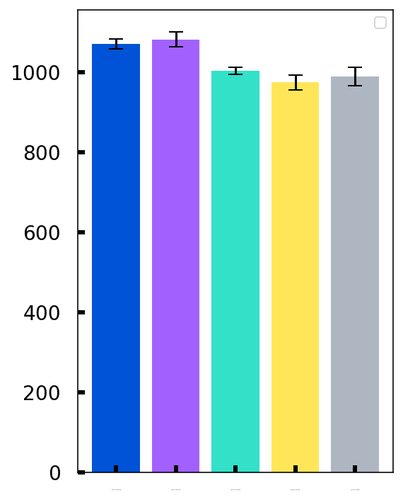}
  \vspace{-0.5em}
  \caption{Mixing ratio with full dataset}  
\label{fig:ablation_data_mix_fulldata}
\end{subfigure}
\begin{subfigure}[b]{0.3\textwidth}
  \centering
  \includegraphics[height=5cm,]{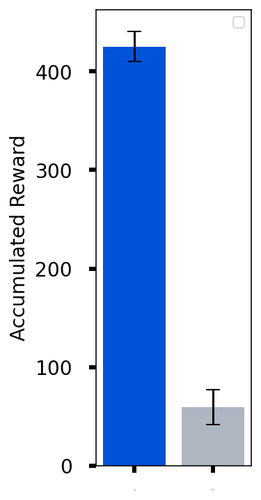}
  \vspace{-0.5em}
  \caption{\met with changing dynamics}  
\label{fig:ablation_dynamics}
\end{subfigure}
\vspace{-0.5em}
\caption{(Left) Comparison with different mixtures of offline and online data when using \met with 4e5 episodes of data on the `Locomotion (State)' task. Accumulated reward (total undiscounted return) for each ratio of data is plotted. A mixture of 50 to 70 percent of online data works best. (Right) Comparison when using \met to reload data to a new task with different dynamics. \met is suprisingly robust and learns to perform better than learning from scratch indicating it's robustness to changes in dynamics.}
\end{figure*}

\rebuttal{
In this section we analyze the effect of reusing \met in a regime where the underlying dynamics change. We reuse the data collected in the `Locomotion Soccer (State)' task but transfer to an environment where a mass is randomly attached to the left and right legs of the walker to perturb it's motion. Figure \ref{fig:ablation_dynamics} shows that \met continues to show an advantage as compared to learning from scratch even when the underlying dynamics of the task have changed. This somewhat surprising finding may be explained by modeling the environment as a partially observed MDP where under some conditions unknown to the agent, the dynamics of walking alter. Reusing data can still guide learning in such a setting showing the robustness of our approach.
}

\end{document}